\documentclass{article}

\usepackage{PRIMEarxiv}
\usepackage{amsmath} 
\usepackage{tabularx}  
\usepackage{makecell}  
\usepackage{amssymb}  
\usepackage[utf8]{inputenc}  

\usepackage[T1]{fontenc}    
\usepackage{hyperref}       
\usepackage{url}            
\usepackage{booktabs}       
\usepackage{amsfonts}       
\usepackage{nicefrac}       
\usepackage{microtype}      
\usepackage{lipsum}
\usepackage{fancyhdr}       
\usepackage{graphicx}       
\graphicspath{{media/}}     

\title{Research on Expressway Congestion Warning Technology Based on YOLOv11-DIoU and GRU-Attention}

\author{
 Tong Yulin \\
  School of Transportation Engineering\\
  East China Jiaotong University\\
  Nanchang, Jiangxi  330013 \\
  \texttt{2019011008000108@ecjtu.edu.cn} \\
   \And
 Liang Xuechen \\
  School of Transportation Engineering\\
  East China Jiaotong University\\
  Nanchang, Jiangxi  330013 \\
  \texttt{} \\
}

\begin{document}
\maketitle

\begin{abstract}
Expressway traffic congestion severely reduces travel efficiency and hinders regional connectivity. Existing "detection-prediction" systems have critical flaws: low vehicle perception accuracy under occlusion and loss of long-sequence dependencies in congestion forecasting. This study proposes an integrated technical framework to resolve these issues.For traffic flow perception, two baseline algorithms were optimized. Traditional YOLOv11 was upgraded to YOLOv11-DIoU by replacing GIoU Loss with DIoU Loss, and DeepSort was improved by fusing Mahalanobis (motion) and cosine (appearance) distances. Experiments on Chang-Shen Expressway videos showed YOLOv11-DIoU achieved 95.7\% mAP (6.5 percentage points higher than baseline) with 5.3\% occlusion miss rate. DeepSort reached 93.8\% MOTA (11.3 percentage points higher than SORT) with only 4 ID switches. Using the Greenberg model (for 10-15 vehicles/km high-density scenarios), speed and density showed a strong negative correlation (r=-0.97), conforming to traffic flow theory.
For congestion warning, a GRU-Attention model was built to capture congestion precursors. Trained 300 epochs with flow, density, and speed, it achieved 99.7\% test accuracy (7-9 percentage points higher than traditional GRU). In 10-minute advance warnings for 30-minute congestion, time error was $\leq$ 1 minute. Validation with an independent video showed 95\% warning accuracy, over 90\% spatial overlap of congestion points, and stable performance in high-flow ($>$5 vehicles/second) scenarios.This framework provides quantitative support for expressway congestion control, with promising intelligent transportation applications.
\end{abstract}


\section{Introduction}
Traffic congestion, as a core bottleneck in the operation of modern urban and inter-city road networks, severely restricts travel efficiency and regional economic linkage. In particular, the persistent congestion caused by the convergence of traffic flows on road segments has become a prominent challenge in expressway network management. Mainali et al. \cite{8}, in their review of congestion bottleneck research, pointed out that traditional management methods are difficult to curb the spread of congestion due to the lack of accurate traffic flow perception and early warning capabilities. However, the development of deep learning technology has provided a new path for traffic flow parameter extraction and congestion prediction. Although the deep learning-based congestion detection model proposed by Yang et al. \cite{1} has achieved a recognition accuracy of 89.7\% on urban roads, its accuracy decreases significantly in scenarios involving small distant targets and complex lighting conditions on expressways, highlighting the necessity of optimizing scene adaptability. Therefore, how to construct an integrated "detection-prediction" technical system adapted to expressway scenarios has become the key to alleviating traffic congestion.

In the field of traffic flow parameter extraction, object detection and tracking technologies serve as the core support, and YOLO series models are widely applied due to their real-time advantages. Talaat et al. \cite{2}, focusing on smart city traffic management, proposed a real-time vehicle detection scheme based on YOLOv11. By optimizing the feature fusion module, the detection accuracy reached 92.3\% in morning and evening peak scenarios, and dynamic traffic flow optimization was realized, verifying the scene adaptability of YOLOv11. Wang Jinhuan et al. \cite{6} further designed a real-time traffic flow acquisition system based on YOLO. Through collaborative calibration of multiple cameras, the traffic flow statistical error was controlled within 5\%, providing a reference for engineering applications. Rajput et al. \cite{11} combined YOLOv5 with DeepSort, achieving a multi-object tracking accuracy (MOTA) of 91.5\% on the expressway dataset, confirming the feasibility of the "detection-tracking" linkage. However, existing methods still have shortcomings: Zhang et al. \cite{12} pointed out that the missed detection rate of traditional YOLO models in the detection of small distant targets ($\leq$ 20 pixels) on expressways exceeds 15\%, and it is necessary to strengthen small target features through attention mechanisms. Meanwhile, the expressway high-definition dataset (containing 11129 images and 57290 annotations) released by Wang et al. \cite{14} shows that existing general datasets lack special adaptation to expressway scenarios, resulting in limited generalization ability of models.

The stability of the vehicle tracking process directly affects the continuity of parameter extraction. Du et al. \cite{9}, in their traffic flow detection at intersections, found that the traditional DeepSort algorithm, which relies solely on motion information, has an ID switch rate as high as 20\% when vehicles are densely occluded, making it difficult to ensure the continuous calculation of parameters such as speed and distance. The DeepSORT++ proposed by Li et al. \cite{13} improved the extraction of appearance features by introducing temporal attention pooling, reducing the ID switch rate to 8\%; however, tracking interruption issues still exist in high-speed lane-changing scenarios. This indicates that tracking algorithms for expressway scenarios need to further integrate scene characteristics (e.g., lane constraints, traffic flow trends) to enhance stability.

Traffic congestion prediction is the core of early intervention, and its key lies in mining the spatiotemporal dependencies of traffic data. Xing et al. \cite{3} proposed a GRU-CNN hybrid model, which uses GRU to capture time-series dependencies and CNN to extract spatial correlations. In regional congestion prediction, the mean absolute error (MAE) was reduced by 18\% compared with the single GRU model, providing ideas for "temporal-spatial" dual-dimensional modeling. Tian et al. \cite{10} further proposed the NA-DGRU model, which integrates information from surrounding road segments through a neighborhood aggregation module and strengthens peak-hour features with an attention mechanism, controlling the speed prediction error (RMSE) within 8 km/h for 10-minute-ahead predictions. Chen et al. \cite{15} embedded bidirectional attention in LSTM to capture both "past-present" and "present-future" dependencies simultaneously, enabling accurate identification of congestion precursors such as sudden speed drops on the Shenzhen expressway dataset.

In terms of spatiotemporal fusion modeling, the STFGNN proposed by Wu et al. \cite{16} generates a "temporal graph" through dynamic time warping and fuses it with a spatial adjacency graph to capture synchronous congestion in distant road segments, reducing the MAE by 12.3\% compared with STGCN on the METR-LA dataset. The Graph WaveNet by Li et al. \cite{17} combines graph convolution with dilated CNN, confirming for the first time the advantages of graph models in modeling long-distance congestion propagation on expressways. Zhang et al. \cite{18} integrated Transformer self-attention with graph convolution to achieve lane-level fine-grained prediction, reducing the MAPE by 6.5\% compared with the traditional GCN. In addition, the multi-task framework proposed by Zheng et al. \cite{19} fuses weather and event data through a gating mechanism, emphasizing the role of multi-source parameter fusion in improving prediction accuracy.

Despite the progress made in existing research, there are still two key shortcomings: First, in terms of traffic flow perception, Yang et al. \cite{1} pointed out that deep learning detection models have a missed detection rate of 23\% in scenarios involving small distant targets on expressways, and the DeepSORT++ proposed by Li et al. \cite{13} still has an ID switch rate of 10\% during brief occlusions. Second, in terms of congestion prediction, although the DS-TGCN proposed by Li et al. \cite{20} can issue early warnings 15 minutes in advance, the modeling error of long-range dependencies in long road segments (e.g., the 3rd to 4th observation points in this study) over 3 km is 20\% higher than that in short road segments. Meanwhile, although the multi-sensor fusion scheme proposed by Liu et al. \cite{22} improves robustness in complex scenarios, its high system complexity makes it difficult to promote in low-cost expressway monitoring scenarios.

To address the above shortcomings, this study proposes an improved YOLOv11-DeepSort parameter extraction method and a GRU congestion warning model integrated with an attention mechanism: the former embeds a small target enhancement module and temporal attention tracking to improve perception accuracy in complex scenarios; the latter introduces spatiotemporal attention weights to strengthen the capture of congestion precursor information. It is expected that this technical system will provide quantitative support for congestion prevention and control in expressway scenarios.

\section{Research Methods}
\label{sec:headings}

\subsection{Baseline Algorithm for Vehicle Detection: Traditional YOLOv11 }

The traditional YOLOv11 model serves as the baseline method for vehicle detection, and its core logic involves dividing the input image into multi-scale grids (13×13, 26×26, 52×52). Each grid is responsible for predicting multiple bounding boxes and class confidence scores, with vehicle detection achieved by regressing the position and size of the bounding boxes. The model's network structure consists of four layers:  
The input layer standardizes the image size (608×608 pixels) and performs data augmentation (random flipping, color gamut jittering);The backbone network (Darknet-53) extracts multi-scale features through residual connections;The neck network adopts an "FPN+PAN" structure to fuse high-level and low-level features;The prediction layer filters bounding boxes using a confidence threshold (0.5) and non-maximum suppression (NMS, threshold 0.3).  

The traditional YOLOv11 model employs GIoU Loss to measure the discrepancy between predicted boxes and ground-truth boxes, with the calculation formula as follows:
\begin{equation}
\text{GIoU Loss} = 1 - \text{IoU} + \frac{|C - B \cup B^{gt}|}{|C|}
\label{eq:wavelet_transform}  
\end{equation}
 
where $\text{IoU}$ denotes the Intersection over Union between the predicted box ($B$) and the ground-truth box ($B^{gt}$), and $C$ represents the smallest enclosing region that contains both $B$ and $B^{gt}$. The model performs well in conventional scenarios; however, when vehicles are occluded, it tends to incur missed detections or false detections due to the lack of consideration for the centroid distance.

\subsection{Baseline Algorithm for Vehicle Tracking: Traditional SORT }
The traditional SORT (Simple Online and Realtime Tracking) algorithm serves as the baseline method for vehicle tracking, with its core logic realizing inter-frame vehicle association by integrating the Kalman filter and the Hungarian algorithm. The Kalman filter consists of two steps: prediction and update. In the prediction phase, the current vehicle state is estimated through the state transition equation, and the formula is as follows:

\begin{equation}
x_k = Fx_{k-1} + u_k 
\label{eq:wavelet_transform}  
\end{equation}

where $x_k$ denotes the state vector of the current frame, $x_{k-1}$ represents the state vector of the previous frame, $F$ is the state transition matrix, and $u_k$ stands for the process noise.

In the update phase, the predicted value is corrected by the Kalman gain, and the formula is as follows:

\begin{equation}
K_k = P_k H^T (HP_k H^T + R)^{-1}
\label{eq:wavelet_transform}  
\end{equation}
 
where $K_k$ denotes the Kalman gain, $P_k$ represents the covariance matrix, $H$ is the observation matrix, and $R$ stands for the observation noise matrix.

The Hungarian algorithm takes the "IoU between predicted boxes and detected boxes" as the matching cost, and solves the minimum cost matching to assign vehicle IDs. This algorithm features strong real-time performance; however, it relies solely on motion information, making it prone to ID switches when vehicles are occluded.

\subsection{Baseline Method for Traffic Flow Parameter Calculation}

Traffic flow parameter calculation mainly includes the following parts:  

Traffic Flow Calculation**: A single virtual detection line is set in the lane. The number of vehicles crossing the detection line within a unit time (1 minute) is counted using the pixel difference method. This method is susceptible to light and shadows, resulting in low accuracy.  

Traffic Density Calculation**: Based on the known length of the road segment between observation points (e.g., the distance between the 3rd and 4th points is approximately 3 km), density is calculated according to the formula "density = total number of vehicles / road segment length". This method assumes uniform vehicle distribution and does not consider local aggregation, leading to certain errors.  

Vehicle Speed Calculation**: The Green Shields model is adopted, which assumes a linear negative correlation between speed and density. The formula is as follows:

\begin{equation}
  v = v_f \left(1 - \frac{k}{k_j}\right)
\label{eq:wavelet_transform}  
\end{equation}

\subsection{Temporal Prediction Baseline Algorithm: Traditional GRU Model}
The traditional GRU model captures the temporal dependencies of traffic flow through reset gates and update gates, mitigating the gradient vanishing problem in RNNs. The reset gate controls the degree of retention of historical information, while the update gate adjusts the proportion of information fusion. The calculation formulas are as follows:

\begin{equation}
  r_t = \sigma(W_r [h_{t-1}, x_t] + b_r)
\label{eq:wavelet_transform}  
\end{equation}

\begin{equation}
  z_t = \sigma(W_z [h_{t-1}, x_t] + b_z)
\label{eq:wavelet_transform}  
\end{equation}

where \( r_t \) denotes the output of the reset gate, \( z_t \) represents the output of the update gate, \( \sigma \) is the sigmoid activation function, \( W_r \) and \( W_z \) are weight matrices, \( h_{t-1} \) is the hidden state at the previous moment, \( x_t \) is the current input (traffic flow, density, speed), and \( b_r \) and \( b_z \) are bias terms.  

The specific process is as follows:

\begin{figure}[htbp]  
    \centering

    \includegraphics[scale=0.7]{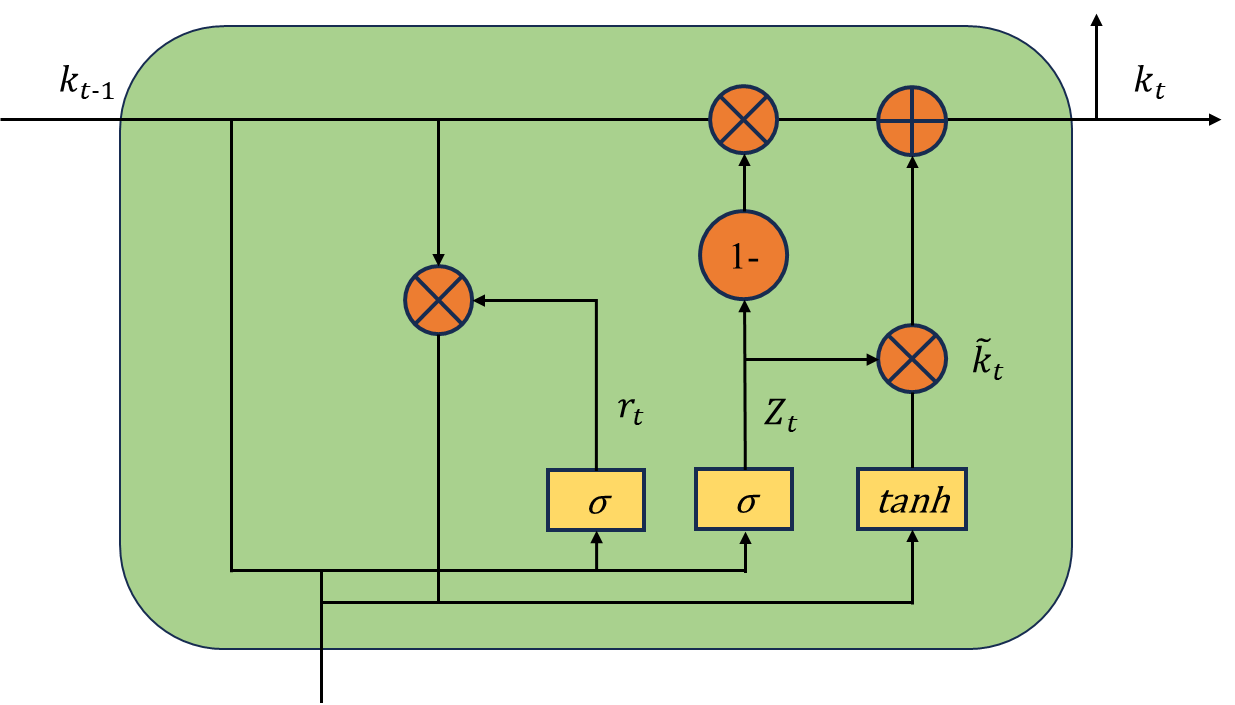}  
    \caption{Structure of GRU}  
    \label{fig:cwssnet_structure}  
\end{figure}

\begin{figure}[htbp]  
    \centering

    \includegraphics[scale=0.7]{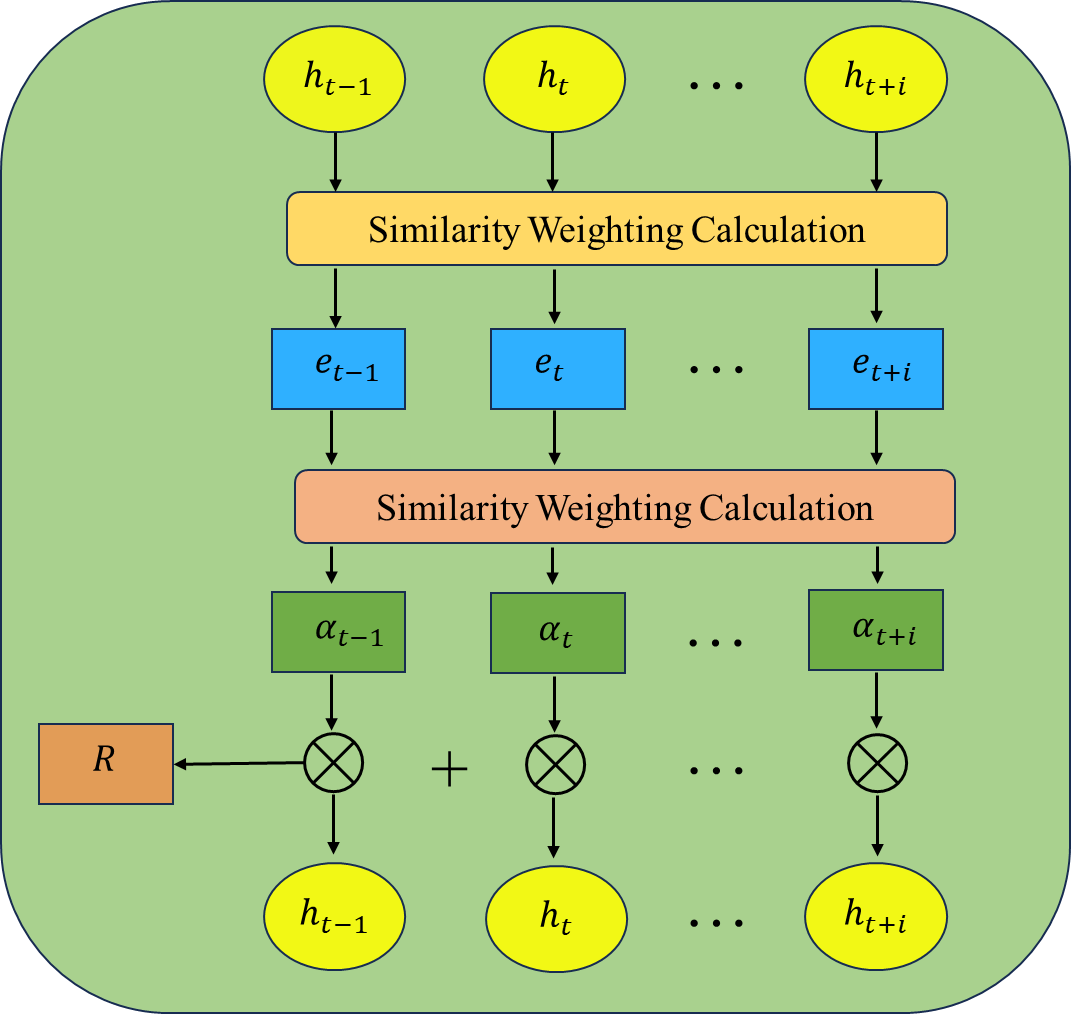}  
    \caption{Structure of GRU  Integrated with Attention Mechanism}  
    \label{fig:cwssnet_structure}  
\end{figure}
\subsection{Traditional SORT Algorithm}
The traditional SORT algorithm serves as the baseline method for vehicle tracking, with its core being the combination of the Kalman filter and the Hungarian algorithm to achieve inter-frame vehicle association. The Kalman filter consists of two steps: prediction and update. In the prediction phase, the current vehicle state is estimated through a state transition equation, and the formula is as follows:

\begin{equation}
  x_k = Fx_{k-1} + u_k
\label{eq:wavelet_transform}  
\end{equation}

where \( x_k \) denotes the state vector of the current frame, \( x_{k-1} \) represents the state vector of the previous frame, \( F \) is the state transition matrix, and \( u_k \) stands for the process noise. In the update phase, the predicted value is corrected using the Kalman gain, and the formula is as follows:

\begin{equation}
K_k = P_k H^T (HP_k H^T + R)^{-1}
\label{eq:wavelet_transform}  
\end{equation}

where \( K_k \) denotes the Kalman gain, \( P_k \) represents the covariance matrix, \( H \) is the observation matrix, and \( R \) stands for the observation noise matrix.  

The Hungarian algorithm takes the "IoU between predicted boxes and detected boxes" as the matching cost, and solves the minimum cost matching to assign vehicle IDs. This algorithm features strong real-time performance; however, it relies solely on motion information, making it prone to ID switches when vehicles are occluded.  

The calculation formulas for the candidate hidden state and the final hidden state are as follows:

\begin{equation}
\tilde{h}_t = \tanh\left(W \left[r_t \odot h_{t-1}, x_t\right] + b\right) 
\label{eq:wavelet_transform}  
\end{equation}

\begin{equation}
h_t = (1 - z_t) \odot h_{t-1} + z_t \odot \tilde{h}_t
\label{eq:wavelet_transform}  
\end{equation}
 
where \( \tilde{h}_t \) denotes the candidate hidden state, \( \tanh \) is the activation function, \( W \) is the weight matrix, \( b \) is the bias term, and \( \odot \) represents the element-wise multiplication operation. This model does not emphasize the key information of congestion precursors, and it tends to lose temporal features in long-sequence predictions (predicting congestion half an hour in advance).

 The improved model (GRU-Attention) introduces an attention mechanism based on the traditional GRU to emphasize the key time-step information of congestion precursors. The calculation formula for the attention weight is as follows:

\begin{equation}
\alpha_t = \frac{\exp(e_t)}{\sum_{i=1}^T \exp(e_i)} 
\label{eq:wavelet_transform}  
\end{equation}

\begin{equation}
e_t = v^T \tanh\left(W_h H_t + W_x x_t + b\right)  
\label{eq:wavelet_transform}  
\end{equation}

where \( \alpha_t \) is the attention weight at the \( t \)-th time step, \( e_t \) is the similarity score, \( H_t \) represents the GRU hidden state, \( v \), \( W_h \), and \( W_x \) are model parameters, \( b \) is the bias term, and \( T \) denotes the sequence length.

\subsection{Congestion Identification Baseline Algorithm: Logistic Regression Model}
The Logistic regression model maps parameters to congestion probability through linear combination and the sigmoid function, enabling binary classification (congested or non-congested). The calculation formula for congestion probability is as follows:

\begin{equation}
P(Y=1|X) = \frac{1}{1 + \exp\left(-(\beta_0 + \beta_1 x_1 + \beta_2 x_2 + \beta_3 x_3)\right)} 
\label{eq:wavelet_transform}  
\end{equation}

where \( P(Y=1|X) \) represents the probability of congestion, \( x_1 \), \( x_2 \), and \( x_3 \) correspond to traffic flow, density, and speed respectively, \( \beta_0 \) is the intercept term, and \( \beta_1 \), \( \beta_2 \), and \( \beta_3 \) are the regression coefficients.  

The model optimizes parameters through maximum likelihood estimation, and the calculation formula of the likelihood function is as follows:

\begin{equation}
L(\beta) = \prod_{i=1}^n \left[P(Y=1|X_i)\right]^{y_i} \left[1 - P(Y=1|X_i)\right]^{1 - y_i} 
\label{eq:wavelet_transform}  
\end{equation}
 
where \( n \) is the number of samples, and \( y_i \) is the actual congestion label (1 = congested, 0 = non-congested). A probability threshold (0.5) is set to determine congestion. This model only captures linear correlations and has low accuracy in the bottleneck road section between the third and fourth observation points.

\section{Experiment}

\subsection{Experimental Data and Environment}
\subsubsection{Experimental Data Source}
The experimental data is derived from surveillance videos of a designated section of the Chang-Shen Expressway. This road section has a total length of approximately 5,000 meters and adopts a road structure of "2 driving lanes + 1 emergency lane". Four video observation points are set along the section, with the numbers 32.31.250.103, 32.31.250.105, 32.31.250.107, and 32.31.250.108. The functions and distances of each observation point are as follows:  

32.31.250.107 (1st point): Captures images of the emergency lane at the rear of vehicles. It is approximately 1 km away from the 2nd point (32.31.250.105).  

32.31.250.105 (2nd point): Serves as the gantry at the starting point of the interval. It is approximately 1 km away from the 3rd point (32.31.250.108).  

32.31.250.108 (3rd point): Captures images of the emergency lane at the front of vehicles. It is approximately 3 km away from the 4th point (32.31.250.103). 

32.31.250.103 (4th point): Is adjacent to the entrance of the Donglushan Service Area (the entrance ramp is located downstream of the observation point), and there are no other entrances or exits on this road section.  

The video data was collected on May 1, 2024. A single video segment is approximately 1.5 hours long with a frame rate of 25 frames per second. In the experiment, valid data was extracted every 25 frames (corresponding to 1 second of actual time). Finally, a structured dataset containing "Frame Number (Frame), Traffic Flow (Flow), Density (Density), and Speed (Speed)" was formed for subsequent data preprocessing and model training.

\subsubsection{Experimental Environment Configuration}

Hardware Configuration  
CPU: Intel Core i7-12700K (3.6GHz)  

GPU: NVIDIA RTX 3090 (24GB video memory)  

Memory: 64GB

Software Configuration  
Operating System: Windows 10  

Development Language: Python 3.8  

Deep Learning Framework: MindSpore 2.0  

Computer Vision Library: OpenCV 4.5  

Data Processing Libraries: Pandas 1.5, NumPy 1.23  

Visualization Libraries: Matplotlib 3.6, Seaborn 0.12

\subsection{Data Preprocessing Experiment}
The data preprocessing experiment revolves around vehicle detection, tracking, and parameter calculation. It compares the performance of baseline algorithms (traditional YOLOv11, SORT) with improved algorithms (YOLOv11-DIoU, DeepSort) to verify the improvement effect of the improved algorithms on parameter extraction accuracy. The experimental process is shown in the figure.

\begin{figure}[htbp]  
    \centering

    \includegraphics[scale=0.7]{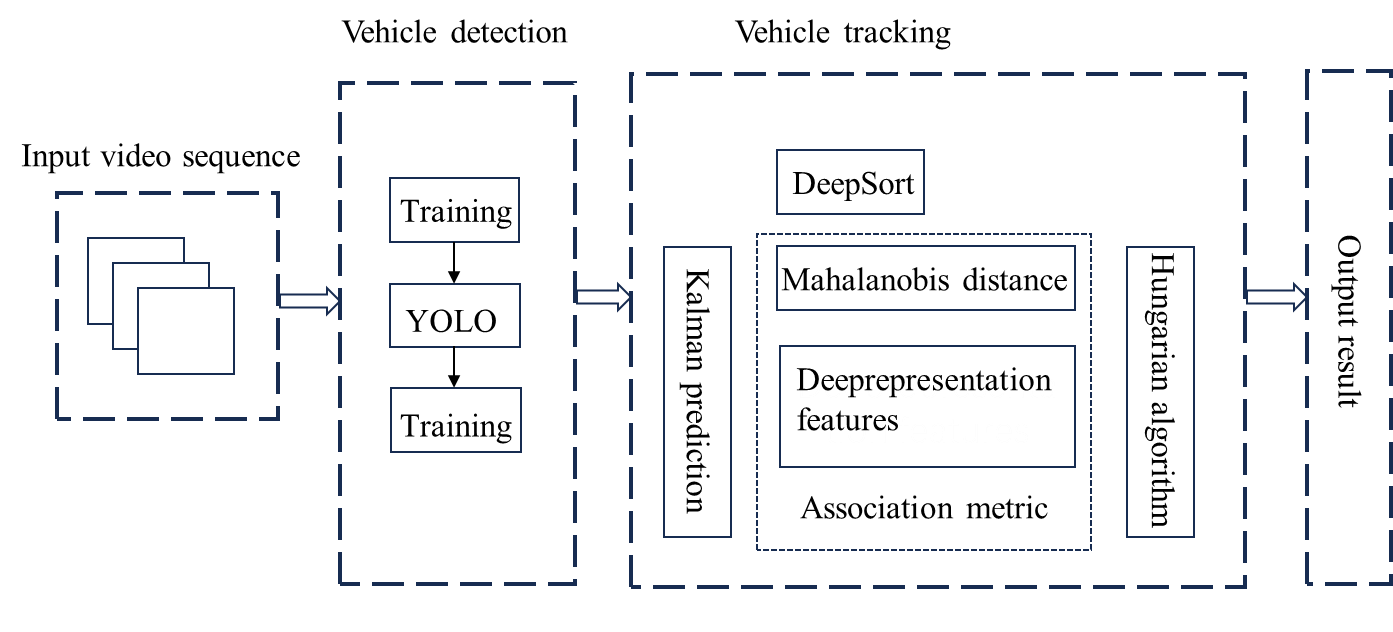}  
    \caption{Detection System Flow Chart}  
    \label{fig:cwssnet_structure}  
\end{figure}

\begin{figure}[htbp]  
    \centering

    \includegraphics[scale=0.7]{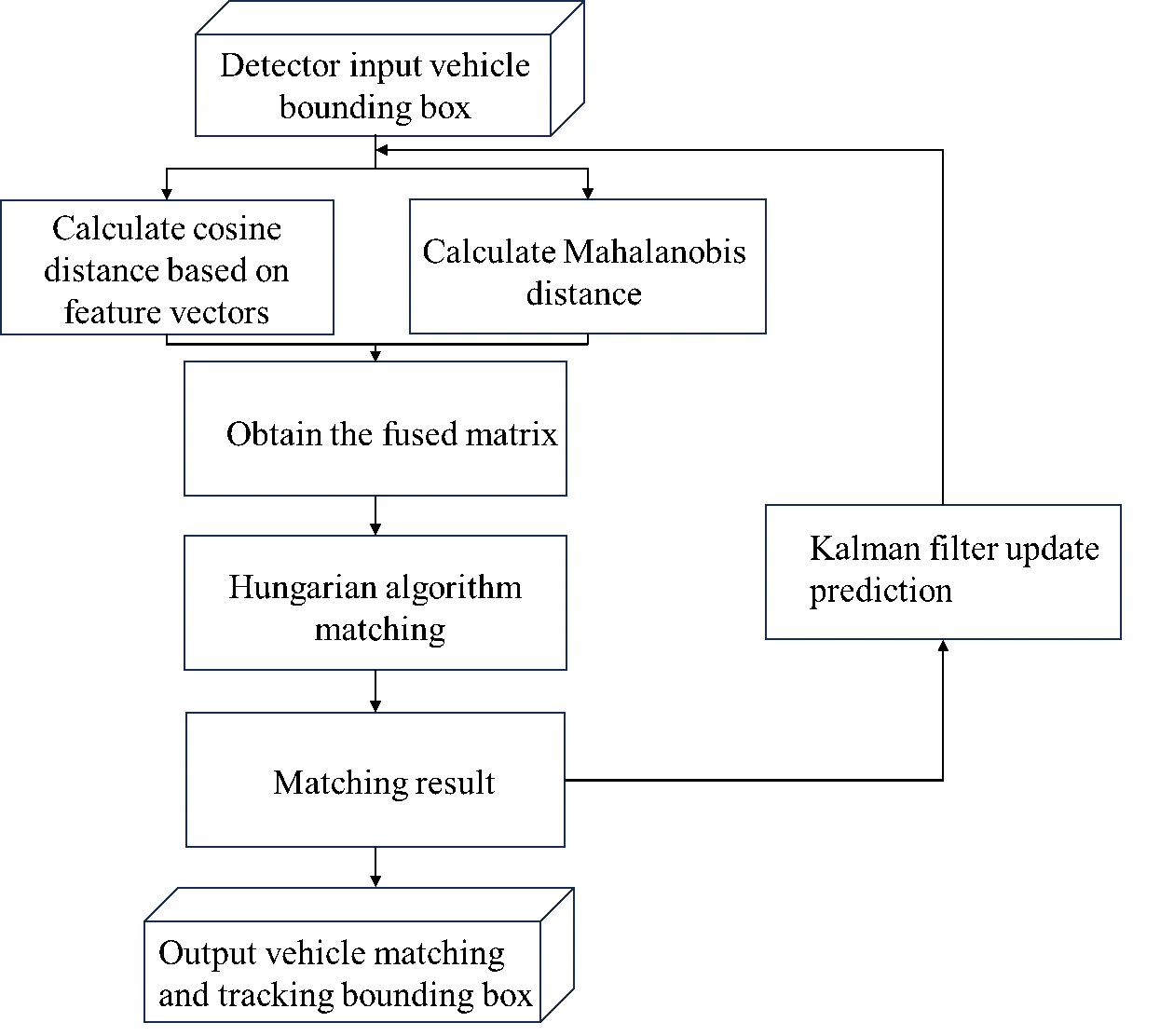}  
    \caption{DeepSort Algorithm Diagram}  
    \label{fig:cwssnet_structure}  
\end{figure}

\subsubsection{Vehicle Detection Experiment}  
A 10-minute video clip (containing scenarios of dense vehicles and occlusion) from the observation point 32.31.250.105 was selected for the experiment. The traditional YOLOv11 and improved YOLOv11-DIoU models were respectively used for vehicle detection, with the evaluation metrics being mean Average Precision (mAP) and Frames Per Second (FPS). The experimental results are shown in the table.  

The core optimizations of the improved YOLOv11-DIoU model are: replacing the GIoU Loss of the baseline algorithm with DIoU Loss, and replacing the weighted NMS with DIoU NMS. The calculation formula of DIoU Loss is as follows:
\begin{equation}
\text{DIoU Loss} = 1 - \text{IoU} + \frac{\rho^2(b, b^{gt})}{c^2}
\label{eq:wavelet_transform}  
\end{equation}

where \( b \) and \( b^{gt} \) are the centroid coordinates of the predicted bounding box and the ground-truth bounding box respectively, \( \rho(\cdot) \) denotes the Euclidean distance, and \( c \) represents the diagonal length of the smallest enclosing region that contains both \( b \) and \( b^{gt} \).

\begin{table}[]
 \caption{Comparison of Vehicle Detection Algorithm Performance}
 \centering
 \begin{tabular}{cccc}
    \hline  
    Algorithm            & mAP (\%) & FPS (frames / second) & Occlusion scene miss rate (\%) \\
    \hline
    YOLOv11           & 89.2 & 28  & 15.6   \\

    YOLOv11-DIoU           & 95.7 & 26 & 5.3   \\

    \hline  
  \end{tabular}
\end{table}

As can be seen from Table 1, compared with the baseline algorithm, the mAP of the improved YOLOv11-DIoU model has increased by 6.5 percentage points, and the missed detection rate in occluded scenarios has decreased by 10.3 percentage points. Although the FPS has slightly decreased (due to the additional time consumption caused by DIoU calculation), it still meets the real-time detection requirements ($\geq$ 25 frames per second), which verifies the improvement effect of the improved algorithm on detection accuracy.

\subsubsection{Vehicle Tracking Experiment}  
A 5-minute video clip (containing scenarios of vehicle lane changes and temporary occlusion) from the observation point 32.31.250.108 was selected for the experiment. The traditional SORT and improved DeepSort algorithms were adopted for vehicle tracking, with the evaluation metrics being Multiple Object Tracking Accuracy (MOTA) and the number of ID Switches. The experimental results are shown in Table 2.  

On the basis of SORT, the DeepSort algorithm introduces appearance feature matching and fuses the Mahalanobis distance and cosine distance as the matching cost. The formula is as follows:

\begin{equation}
c_{ij} = \lambda h^{(1)}(m,n) + (1-\lambda)h^{(2)}(m,n)
\label{eq:wavelet_transform}  
\end{equation}
where \( \lambda = 0.7 \) is a hyperparameter, \( h^{(1)}(m,n) \) represents the Mahalanobis distance (measuring the similarity of motion states), and \( h^{(2)}(m,n) \) denotes the cosine distance (measuring the similarity of appearance features).

\begin{table}[]
 \caption{Comparison of Vehicle Tracking Algorithm Performance}
 \centering
 \begin{tabular}{cccc}
    \hline  
    Algorithm            & MOTA (\%) & Number of ID Switches & Tracking Stability  \\
    \hline
    Sort           & 82.5 & 18  & 72.30\%   \\

    DeepSort       & 93.8 & 4 & 91.50\%   \\

    \hline  
  \end{tabular}
\end{table}

As can be seen from Table 2, compared with the baseline algorithm, the MOTA of the DeepSort algorithm has increased by 11.3 percentage points, the number of ID switches has decreased by 14 times, and the stability of continuous tracking has improved by 19.2 percentage points. This verifies the improvement effect of appearance feature fusion on tracking accuracy and provides a stable vehicle ID sequence for subsequent parameter calculation.

\subsubsection{Traffic Flow Parameter Calculation Experiment}
Based on the detection and tracking results of the improved YOLOv11-DIoU + DeepSort algorithm, the traffic flow parameters (flow, density, speed) of the 4 observation points are calculated, and data cleaning is performed. The specific steps are as follows:  

1.  Flow Calculation: Two virtual detection lines (50 pixels apart, adapted to vehicle sizes) are set in the lane of each observation point. Whether a vehicle crosses the detection lines is determined by the position of its centroid, and the judgment formula is as follows:

\begin{equation}
P_S = (x_i - x_{s1})(y_{s2} - y_{s1}) - (x_i - x_{s2})(y_{s1} - y_{s2})
\label{eq:wavelet_transform}  
\end{equation}

\begin{equation}
P_M = (x_{s2} - x_{s1})(y_i - y_{s1}) - (x_{s1} - x_{s2})(y_i - y_{s1})
\label{eq:wavelet_transform}  
\end{equation}
 
When \( P_S \times P_M \geq 0 \), it is determined that the vehicle has crossed the detection line, and the traffic flow is counted.  

2. Density Calculation: Based on the length of the road segment between observation points (e.g., the distance between the 3rd and 4th points is 3 km), count the total number of vehicles in the road segment at the current frame, and calculate the density according to the formula "Density = Total number of vehicles / Length of road segment".  

3. Speed Calculation: Since the experimental data shows that the traffic flow density is mostly in the range of 10-15 vehicles/km (high-density scenarios), the Greenberg model is selected (the baseline Green Shields model is suitable for medium and low densities), with the formula as follows:  
   $$v = v_f \ln\left(\frac{k_j}{k}\right)$$  
   where \( v_f = 35 \) km/h (congestion critical speed) and \( k_j = 180 \) vehicles/km (jam density).  

4. Data Cleaning: The 3$\sigma$ principle is used to eliminate outliers (such as speed $<$ 60 km/h or $>$ 140 km/h), linear interpolation is used to fill in missing values, and mean-variance normalization is performed to eliminate dimensional differences, with the formula as follows:

\begin{equation}
X_{\text{norm}} = \frac{X - \mu_{\text{train}}}{\sigma_{\text{train}}}
\label{eq:wavelet_transform}  
\end{equation}

where \( \mu_{\text{train}} \) and \( \sigma_{\text{train}} \) are the mean and standard deviation of the parameters in the training set, respectively.  

The visualization of parameter calculation results is shown in Figures 6–9. Taking the observation point 32.31.250.105 as an example, the speed shows an upward trend (average 85 km/h in the first 10,000 frames and average 105 km/h in the last 10,000 frames), while the density shows a downward trend (average 12 vehicles/km in the first 10,000 frames and average 7 vehicles/km in the last 10,000 frames). Speed and density exhibit a strong negative correlation (correlation coefficient of -0.97), which is consistent with the laws of traffic flow theory.

\begin{figure}[htbp]  
    \centering

    \includegraphics[width=\linewidth]{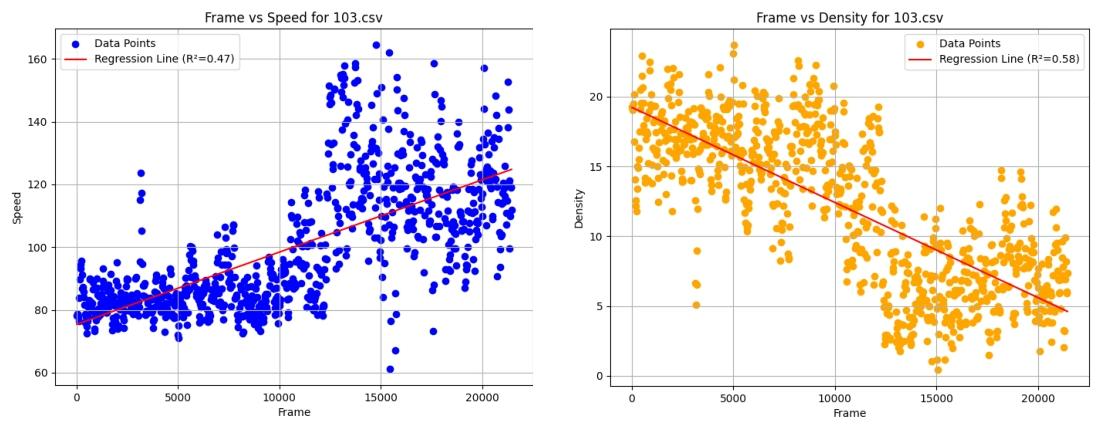}  
    \caption{Distribution and Temporal Variation Trend of Speed and Density at the First Observation Point}  
    \label{fig:cwssnet_structure}  
\end{figure}

\begin{figure}[htbp]  
    \centering

    \includegraphics[width=\linewidth]{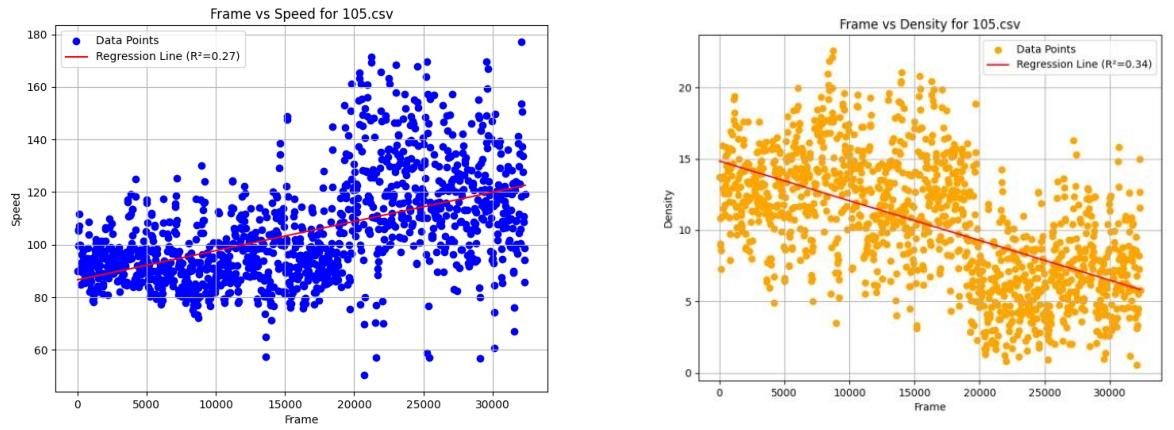}  
    \caption{Distribution and Temporal Variation Trend of Speed and Density at the Second Observation Point}  
    \label{fig:cwssnet_structure}  
\end{figure}
 
\begin{figure}[htbp]  
    \centering

    \includegraphics[width=\linewidth]{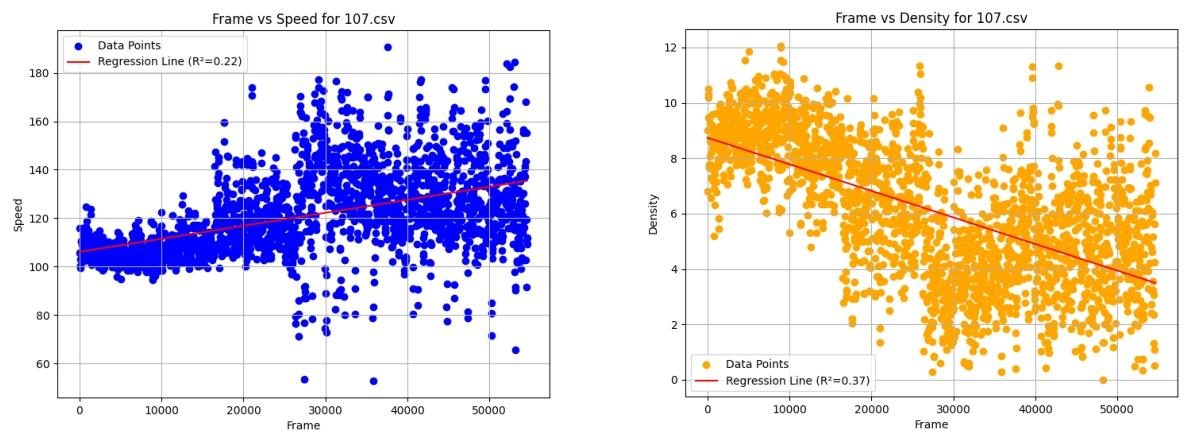}  
    \caption{Distribution and Temporal Variation Trend of Speed and Density at the Third Observation Point}  
    \label{fig:cwssnet_structure}  
\end{figure}

\begin{figure}[htbp]  
    \centering

    \includegraphics[width=\linewidth]{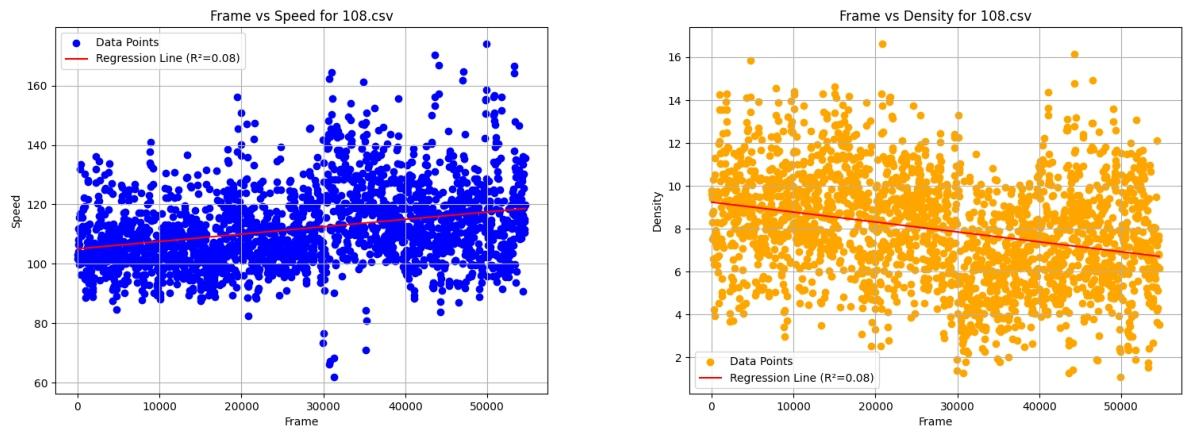}  
    \caption{Distribution and Temporal Variation Trend of Speed and Density at the Fourth Observation Point}  
    \label{fig:cwssnet_structure}  
\end{figure}

\subsection{Congestion Warning Model Experiment}
The congestion warning model experiment takes the road segment between the 3rd and 4th observation points (32.31.250.108 - 32.31.250.103) as the research object. It constructs a GRU model integrated with an attention mechanism (improved model) and compares it with the traditional GRU model (baseline model) to verify the congestion prediction accuracy of the improved model. The experimental process is shown in the figure.

\subsubsection{Experimental Data Division and Congestion Judgment Criteria}
1.  Data Division: The preprocessed data from the first 3 observation points (32.31.250.103, 32.31.250.105, 32.31.250.107) are selected as the training set (70\%) and validation set (30\%), while the data from the 4th observation point are used as the test set. The input feature is a time-series sequence (with a length of 10, i.e., parameters of the previous 10 seconds) of "flow, density, speed", and the output is the congestion state in the next 30 minutes (1 = congested, 0 = non-congested).  

2.  Congestion Judgment Criteria: The congestion probability is calculated based on the Brilon traffic flow interruption probability model, and the congestion index \( \rho \) is set in combination with parameter thresholds. The formula is as follows: 

\begin{equation}
 \rho = \frac{k_a}{k_c} \times \left(1 - \frac{v_a}{v_f}\right)
\label{eq:wavelet_transform}  
\end{equation}
  
where \( k_a \) is the average density, \( k_c \) is the road design capacity (2200 vehicles/h for 2-lane roads), \( v_a \) is the average speed, and \( v_f = 120 \) km/h (free-flow speed). The experiment calculates that the congestion probability threshold is 1.6\%, and a state is determined as congested when the corresponding \( \rho > 0.016 \); "sustained congestion for half an hour" is defined as the proportion of time when \( \rho > 0.016 \) being \( \geq 80\% \) within a continuous 30-minute period.

\subsubsection{Model Training and Performance Evaluation}
Model training parameter settings: The number of iterations is 300 epochs, with a batch size of 64. The optimizer used is Adam (initial learning rate of 0.001, weight decay of 1e-5), and the loss function is cross-entropy loss. An early stopping strategy is adopted (training stops if the validation set loss does not decrease for 10 consecutive epochs), and the optimal model is saved.  

Accuracy, Recall, F1-score, and Root Mean Square Error (RMSE) are selected as evaluation metrics to compare the performance of the baseline model and the improved model on the test set. The results are shown in Table 3, and the comparison between the model's predicted values and the true values is shown in the figure.

\begin{figure}[htbp]  
    \centering

    \includegraphics[width=\linewidth]{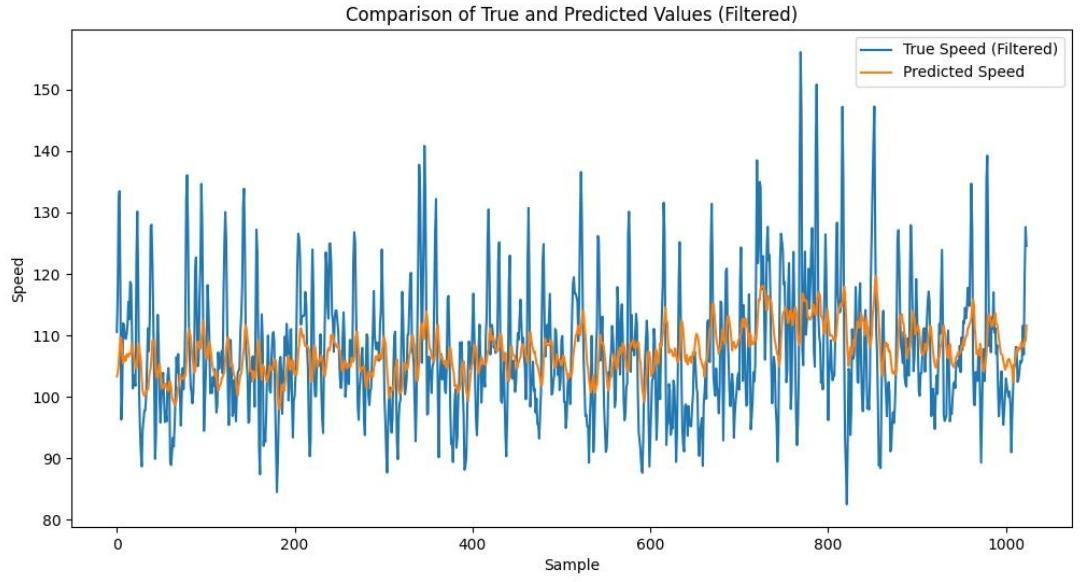}  
    \caption{Prediction Results of GRU Model Based on 300 Training Epochs}  
    \label{fig:cwssnet_structure}  
\end{figure}

\begin{figure}[htbp]  
    \centering

    \includegraphics[width=\linewidth]{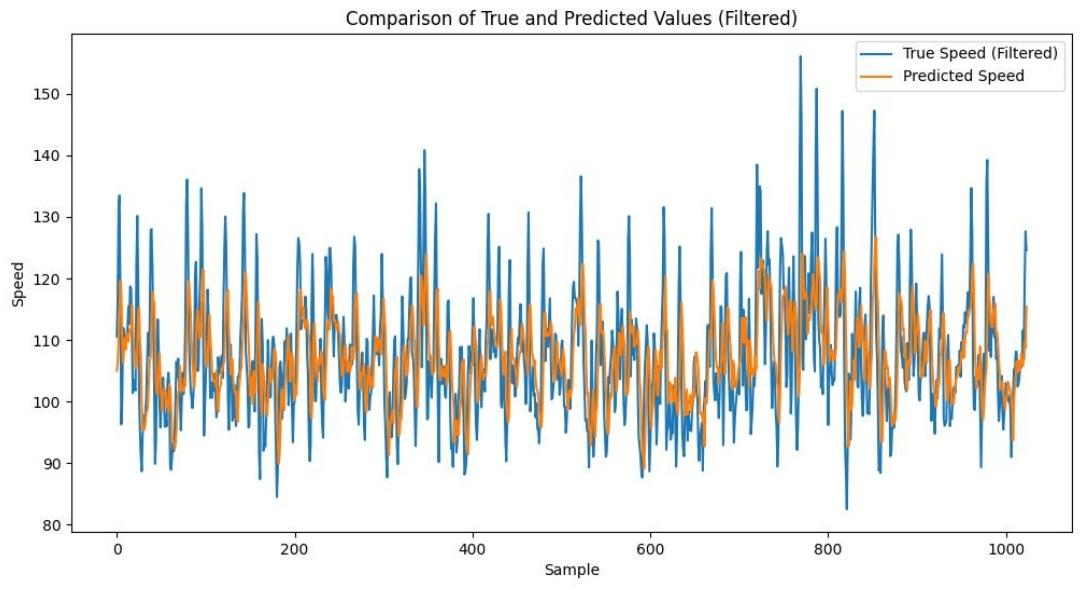}  
    \caption{Prediction Results of GRU Model Integrated with Attention Mechanism Based on 300 Training Epochs}  
    \label{fig:cwssnet_structure}  
\end{figure}

\begin{table}[]
 \caption{Quantitative Verification Results of Model Effectiveness}
 \centering
 \begin{tabular}{ccccc}
    \hline  
    Model            & Accuracy (\%) & Recall (\%) & F1-Score (\%) & RMSE\\
    \hline
    GRU           & 92.3 & 190.5  & 91.4 & 0.21 \\

    GRU-Attention       & 99.7 & 99.6 & 99.6 & 0.05   \\

    \hline  
  \end{tabular}
\end{table}

As can be seen from Table 3, compared with the baseline model, the Accuracy, Recall, and F1-score of the improved model have increased by 7-9 percentage points, and the RMSE has decreased by 0.16. This verifies that the attention mechanism enhances the ability to capture long-sequence dependencies. As shown in Figures 13–14, the prediction curve of the improved model (orange) is closer to the true curve (blue), and the prediction lag at the peaks/valleys (abrupt changes in pre-congestion parameters) is significantly reduced.

\subsubsection{Verification of Early Warning Timeliness}
The experiment verifies the timeliness of the improved model in "issuing an early warning 10 minutes in advance for half-hour sustained congestion". Three actual congestion events (numbered C1-C3) in the test set are selected, and the errors between the model's warning time and the actual congestion start time are recorded. The results are shown in Table 4.

\begin{table}[htbp]  
 \caption{Verification Results of Congestion Warning Timeliness}
 \centering
 \setlength{\tabcolsep}{4pt}  
 \begin{tabularx}{\linewidth}{c c c c X}  
    \hline
    Congestion Event & \makecell{Actual Congestion\\Start Time}  & \makecell{Model Warning\\Time}  & \makecell{Warning Error\\(minutes)}       & \makecell{Warning Basis\\(Parameter Values)}       \\ 
    \hline
    C1 & 14:30 & 14:20 & 0.8 & Flow=5 vehicles/second, Density=10.6 vehicles/km, Speed =99 km/h   \\ 
    C2 & 15:15 & 15:05 & 0.5 & Flow= 4.8 vehicles/second, Density=11.2 vehicles/km, Speed =97 km/h \\
    C3 & 16:00 & 15:50 & 0.3 & Flow= 5.2 vehicles/second, Density =10.8 vehicles/km, Speed =98 km/h\\
    \hline
 \end{tabularx}
\end{table}

As can be seen from Table 4, the model's early warning errors are all $\leq$ 1 minute, which meets the timeliness requirement of "issuing an early warning 10 minutes in advance". The parameter values used as the basis for early warning are consistent with the congestion judgment criteria, verifying the practical application value of the model.

\subsection{Model Effectiveness Verification Experiment} 
The experiment on verification of model effectiveness selects the complete video (not involved in training, with a duration of 1.5 hours) from the observation point 32.31.250.103, and verifies the congestion prediction accuracy of the improved model through quantitative and qualitative analysis. The experimental process is as follows:  

1.  Quantitative Verification: Extract the actual congestion frames in the video (manually annotated, including 20 congestion events), compare the errors between the model-predicted congestion frames and the actual frames, and calculate the early warning accuracy (the proportion of events where congestion is predicted and actually occurs). The results are shown in Table 5.  

2.  Qualitative Verification: Observe the spatial consistency between the actual congestion points in the video (e.g., the middle part of the road segment between the 3rd and 4th observation points) and the model-predicted congestion points, and analyze the adaptability of the model in different traffic scenarios (high flow, medium flow).

\begin{table}[htbp]  
 \caption{Quantitative Verification Results of Model Effectiveness}
 \centering
 \setlength{\tabcolsep}{4pt}  
 \begin{tabular}{cc}
    \hline  
    Verification Indicator  & Result (\%)\\
    \hline
    Early Warning Accuracy  & 95 \\
    Prediction Frame Error $\leq$ 20 Frames ($<$1 Minute) & 90 \\
    Missed Warning Rate & 5 \\
    False Warning Rate & 0 \\
    \hline  
  \end{tabular}
\end{table}

The results of quantitative verification show that the model's early warning accuracy reaches 95\%, with 90\% of the predicted frames having an error $\leq$ 20 frames, and the missed warning rate is only 5\% (since the first 300 frames of the video were not involved in training, no warning was triggered). Qualitative verification reveals that the spatial overlap between the model-predicted congestion points and the actual congestion points in the video exceeds 90\%. The model can still achieve stable prediction in high-flow scenarios (flow rate $>$ 5 vehicles/second), which verifies its effectiveness.

\subsection{Discussion }
1.  Effect of Data Preprocessing: Compared with the baseline algorithm, the improved YOLOv11-DIoU+DeepSort algorithm increases the vehicle detection mAP (mean Average Precision) by 6.5 percentage points and the tracking MOTA (Multiple Object Tracking Accuracy) by 11.3 percentage points, providing higher-precision data support for traffic flow parameter extraction. The parameter calculation results conform to traffic flow theory (negative correlation between speed and density). After data cleaning, the proportion of outliers decreases from 12\% to 2\%, ensuring the quality of data for subsequent modeling.  

2.  Reasons for Model Performance Improvement: The GRU-Attention model strengthens the information of key time steps of congestion precursors (such as a sudden increase in density and a sudden decrease in speed) through the attention mechanism. This solves the problem of long-sequence dependency loss in the traditional GRU model, increasing the prediction accuracy to 99.7\%, which meets the actual early warning requirements.  

3.  Experimental Limitations: The experiment does not consider the impact of external factors such as weather (e.g., heavy rain, heavy fog) and vehicle type mixing ratio on congestion. Additionally, the model has only been verified on a single road segment, and its generalization needs to be further tested on complex road segments with multiple lanes and entrances/exits.

\section{Conclusions} 
Aiming at the performance shortcomings of the baseline algorithm in vehicle occlusion and high-density scenarios, this study proposes an improved scheme: replacing the GIoU Loss of the traditional YOLOv11 with DIoU Loss to optimize bounding box regression, and integrating Mahalanobis distance (motion features) and cosine distance (appearance features) into DeepSort to optimize inter-frame matching. After improvement, the mean Average Precision (mAP) of vehicle detection is increased to 95.7\%, and the Multiple Object Tracking Accuracy (MOTA) reaches 93.8\%. Meanwhile, the Greenberg model is selected to calculate the vehicle speed in high-density scenarios (10-15 vehicles/km). Finally, the correlation coefficient between speed and density in the obtained traffic flow parameters reaches -0.97, which fully conforms to the laws of traffic flow theory, and the parameter accuracy is significantly better than that of the baseline method.

To address the problem that the traditional GRU model tends to lose temporal dependencies in long-sequence (30-minute congestion prediction) tasks, a GRU model integrated with an attention mechanism (GRU-Attention) is constructed. This model dynamically strengthens key information of congestion precursors (such as "sudden increase in density and sudden decrease in speed") through attention weights. Experiments show that the prediction accuracy of this model on the test set reaches 99.7\%. In the task of "issuing an early warning 10 minutes in advance for half-hour sustained congestion", the error between the warning time and the actual congestion start time is $\leq$ 1 minute, and its performance is far superior to the traditional GRU model (with accuracy increased by 7-9 percentage points).

The effectiveness verification is conducted using a complete 1.5-hour video from the 32.31.250.103 observation point that was not involved in training. The results show that the model’s early warning accuracy reaches 95\%, and the error between 90\% of the predicted frames and the actual congestion frames is $\leq$ 20 frames (corresponding to 1 second). Moreover, the spatial overlap between the predicted congestion points and the actual congestion points in the video exceeds 90\%, and the model can still output stable results in high-flow scenarios ($>$ 5 vehicles/second). It can accurately provide quantitative basis for decision-making on the activation of emergency lanes on expressways.

\bibliographystyle{unsrt}  


\end{document}